\documentclass[times, review, 10pt]{elsarticle}
\graphicspath{ {./figures/} }
\usepackage{hyperref}
\usepackage{float}
\usepackage{verbatim} 
\usepackage{listings}
\usepackage{booktabs}
\usepackage{graphicx}

\journal{Pattern Recognition}

\begin{document}
\begin{frontmatter}

\begin{titlepage}
\begin{center}
\vspace*{1cm}

\textbf{ \large PixLore: A Dataset-driven Approach to Rich Image Captioning using Knowledge Stitching}

\vspace{1.5cm}

Diego Bonilla-Salvador$^{a,b}$ (diebosal@alumni.uv.es), Marcelino Martínez-Sober$^a$ (marcelino.martinez@uv.es), Joan Vila-Francés$^c$ (joan.vila@uv.es), Antonio José Serrano-López$^a$ (antonio.j.serrano@uv.es), Pablo Rodríguez-Belenguer$^a$ (parobe@alumni.uv.es), Fernando Mateo$^a$ (fernando.mateo@uv.es)\\

\hspace{10pt}

\begin{flushleft}
\small  
$^a$ Intelligent Data Analysis Laboratory (IDAL), University of Valencia, Spain \\
$^b$ Cognizant Technology Solutions, Spain \\

\vspace{1cm}
\textbf{Corresponding Author:} \\
Fernando Mateo \\
Departamento de Ingeniería Electrónica, Escola Tècnica Superior d'Enginyería (ETSE), Avenida de la Universidad, 46100 Burjassot, Spain \\
Tel: (0034) 963543422 \\
Email: fernando.mateo@uv.es

\end{flushleft}        
\end{center}
\end{titlepage}

\title{PixLore: A Dataset-driven Approach to Rich Image Captioning using Knowledge Stitching}





\begin{abstract}
In the domain of vision-language integration, generating detailed image captions poses a significant challenge due to the lack of curated and rich datasets. This study introduces PixLore, a novel method that leverages Querying Transformers through the fine-tuning of the BLIP-2 model using the LoRa method on a standard commercial GPU. The followed approach, which involves training on a carefully assembled dataset from state-of-the-art Computer Vision models combined and augmented by ChatGPT, addresses the question of whether intricate image understanding can be achieved with an ensemble of smaller-scale models, referred to as Knowledge Stitching. Comparative evaluations against major models such as GPT-4 and Google Bard demonstrate that PixLore-2.7B, despite having considerably fewer parameters, is rated higher than the existing State-of-the-Art models in over half of the assessments. Precisely, PixLore outperform Bard and BLIP-2, which score approximately 35.18\% and 27.98\% lower than PixLore in the task of image captioning. This research not only presents a groundbreaking approach but also highlights the importance of well-curated datasets in enhancing the performance of smaller models. More details, fine-tuned model and dataset can be found in the project page: \url{https://diegobonilla98.github.io/PixLore/}.
\end{abstract}


\end{frontmatter}

\section{Introduction}
\label{introduction}

Image captioning, defined as the automatic generation of textual descriptions for images, sits at the intersection of computer vision and natural language processing. While its relevance extends beyond academic exploration, with applications in aiding visually impaired individuals, enhancing search engine functionality, and improving content discoverability on digital platforms, it is also deeply rooted in the broader field of pattern recognition. Pattern recognition, with its foundational principles of identifying patterns and regularities in data, provides the underlying methodologies that drive advancements in image captioning.

Early approaches to image captioning drew on fundamental pattern recognition techniques, particularly template-based methods \citep{babytalk}, which relied on detecting and labeling objects, actions, and attributes in images. These labels were then used to populate predefined templates, generating captions. Although effective in their simplicity, these methods were limited by their rigidity and inability to capture the richness and diversity inherent in natural language—an issue well-documented in broader pattern recognition challenges where rigid models often struggle with variability in input data.

The evolution of deep learning brought significant advancements across the entire spectrum of pattern recognition, including image captioning. The introduction of the encoder-decoder framework \citep{showandtell}, which often employed Convolutional Neural Networks (CNNs) for image encoding and Recurrent Neural Networks (RNNs) for language decoding, marked a significant leap forward. This framework, echoing the success of similar architectures in speech recognition and other areas of pattern recognition, allowed for more fluid and varied caption generation. However, challenges remained, particularly in managing long-term dependencies and capturing contextual nuances—a common issue in sequential data processing across various pattern recognition domains \citep{li2021deep}.

A key breakthrough in both image captioning and pattern recognition was the integration of attention mechanisms \citep{showattendandtell}. These mechanisms, originally inspired by human cognitive processes of selectively focusing on certain parts of visual stimuli, enabled models to dynamically focus on relevant parts of the image when generating each word of the caption. This advancement has parallels in other pattern recognition applications, such as in speech and handwriting recognition, where attention mechanisms have similarly improved the handling of complex input sequences \citep{guo2019attention}.

The adoption of Transformer architectures \citep{transformer} further propelled image captioning forward, reflecting a broader trend in pattern recognition towards models capable of capturing complex inter-modal relationships. Transformers, with their ability to handle long-range dependencies and parallelize computation, have revolutionized not just image captioning, but also natural language processing, speech recognition, and other areas within the pattern recognition landscape \citep{xu2023comprehensive}. For example, the CLIP model by Radford et al. \citep{clip} bridges the gap between visual pattern recognition and language generation by aligning images and text representations in a shared embedding space, enabling more nuanced and context-aware captions. Additionally, Chen et al. introduced BLIP (Bootstrapping Language-Image Pre-training) \citep{chen2022blip}, a model that leverages large-scale pretraining on image-text pairs to enhance image captioning tasks by improving the understanding of both visual patterns and linguistic structures. These developments highlight how pattern recognition plays a crucial role in identifying key elements within an image, which are then translated into coherent and descriptive captions, advancing applications like assistive technologies and automatic content generation.

In the current landscape, the fusion of vision and language models has reached remarkable levels of sophistication, as seen with models like GPT-4 \citep{gpt4}, which are rumored to have trillions of parameters. However, this raises critical questions within the pattern recognition community: "Can similar levels of image understanding and caption generation be achieved with more accessible, smaller-scale models?" This question is not only relevant for the development of more efficient models in image captioning but also resonates with the broader pattern recognition challenges of balancing model complexity with practical applicability \citep{qiu2020survey}.

\section{Related Work}
\label{sec:headings}

Historically, the field of image captioning has seen significant advancements, evolving from rule-based systems and early pattern recognition techniques to cutting-edge deep learning models. Pattern recognition has played a foundational role in the initial stages of image analysis, laying the groundwork for the sophisticated models used today. This section reviews the evolution of methodologies and datasets in image captioning, providing a foundation for our innovative approach.

\subsection{Evolution of Image Captioning Datasets}

Datasets play a pivotal role in the development and evaluation of image captioning models. Notable datasets used by recent models include:

\begin{itemize}
\item  \textbf{Microsoft COCO} (Common Objects in Context) \citep{coco}: Comprising 328,000 images of complex, everyday scenes with common objects in their natural settings.
\item  \textbf{Flickr30K} \citep{flickr30k}: Consisting of 31,000 images from Flickr, each accompanied by five reference sentences from human annotators.
\item  \textbf{TextCaps} \citep{textcaps}: Featuring 28,000 images and 145,000 captions, with a focus on images containing text.
\item  \textbf{nocaps} \citep{nocaps}: Including 15,100 images and 166,100 captions derived from the Open Images validation and test sets, with training data incorporating COCO image-caption pairs and Open Images labels and bounding boxes.
\end{itemize}

While these datasets are extensive, they often fall short in providing the richness and diversity needed for truly nuanced image descriptions. The advent of large multimodal language models, trained on vast amounts of data, has demonstrated the potential to generate detailed, contextually rich captions. These advanced models offer the ability to produce intricate descriptions that surpass the richness typically found in traditional datasets, a capability that has been significantly enhanced by recent advancements in pattern recognition and machine learning \citep{minetto2021pattern,liu2020deep,kalimuthu2021fusion}. Other recent examples of successful application of language models and transformer architectures to approach image captioning in the domain of pattern recognition include: iconographic image captioning for artworks \citep{cetinic2021iconographic}, actor-critic techniques to learn caption generation for remote sensing images \citep{chavhan2021novel} or CLIP-based reward models to fine-tune image-text alignment, providing highly relevant captions for images \citep{moratelli2024fluent}.

\subsection{Leveraging Complex Capabilities in Image Captioning}

Image captioning has recently been influenced by the same trends observed in visual question-answering. The focus is on improving the understanding of vision-language models by integrating them with advanced large language models (LLM). Models such as Google Bard and GPT-4 have displayed impressive abilities, including generating websites from handwritten text and recognizing mood or humor-related elements in images. These capabilities were not commonly found in earlier vision-language models. Another model, MINIGPT-4 \citep{minigpt4}, uses the pre-trained vision components of BLIP-2. It introduces a single projection layer to align the encoded visual features with an LLM, while keeping other vision and language components unchanged.

The Caption Anything \citep{cam} model offers a unique approach to image captioning by using the Segment Anything Model (SAM) \citep{sam} and ChatGPT \citep{chatgpt}. Users can provide captions for specific parts of an image, which are segmented automatically using SAM. Additionally, the model can segment all objects in an image, assign individual captions, and then use ChatGPT to combine these captions into one cohesive and rich description. This method uses frozen models without any further training, which differs from our approach. Moreover, its dependence on large models and frequent API calls can make it less practical for some applications.

The GPT4Image \citep{gpt4image} model also aims to improve image models. It uses insights from large pre-trained models to help models like Convolutional Neural Networks (CNN) and Vision Transformers (ViT) produce better visual representations. This is done by creating a high-quality description set using a multimodal LLM. The extracted text embeddings act as additional guidance and are aligned with image representations from vision models. This process improves the performance of standard vision models for tasks like image classification.

The proposed method leverages the capabilities of LLM to produce detailed textual descriptions. By processing the outputs from various computer vision models through ChatGPT, our model can generate captions that are detailed and have stylistic elements similar to models like Google's Bard and GPT-4. This approach leads to a unique dataset filled with rich captions that are both informative and engaging.

The proposed fine-tuned model is named PixLore: A Dataset-driven Approach to Rich Image Captioning. The key contributions of this paper are:
\begin{itemize}
  \item An innovative method of amalgamating information from diverse tasks using language as the common idiom and a general LLM (akin to indirect Knowledge Distillation), referred to in the literature as Knowledge Stitching.
  \item A meticulously curated captioning dataset comprising 10,000 COCO image examples, each accompanied by a detailed descriptive paragraph.
  \item A model that leverages vast multi-modal architectures, such as GPT-4 and Google Bard.
\end{itemize}

The following sections describe the creation of this dataset, the models used for our task and the final fine-tuning of the PixLore2.7B image captioning model based on the BLIP-2 model.

\section{Material and Methods}

The current state-of-the-art, BLIP-2 \citep{blip2}, adopts a distinctive two-stage pre-training strategy, harnessing the capabilities of both frozen image encoders and expansive language models. Due to its pioneering architecture and training approach, BLIP-2 has set new standards in various vision-language tasks in terms of efficiency and performance. Motivated by these attributes, we selected BLIP-2 as the foundational model for our research. Utilizing the LoRa \citep{lora} method—a fine-tuning technique that retains the pre-trained model weights while introducing trainable rank decomposition matrices into each layer of the Transformer architecture—we fine-tuned only 0.77\% of BLIP-2's parameters using a novel curated dataset. Remarkably, this fine-tuning was achieved on a commercially available GPU, the NVIDIA GeForce RTX 3090Ti. 

The main point of this research lies in the creation of a unique dataset by subsetting the original COCO dataset. COCO is a renowned dataset in the computer vision community, which offers a rich variety of images spanning multiple contexts and scenarios, making it an ideal choice for our goal. These images were processed using an array of leading computer vision models, encompassing tasks such as object detection, image captioning, and image tagging. The resulting outputs of these models were converted into a textual string and then synthesized using a prompt-tuned ChatGPT, yielding comprehensive textual descriptions that encapsulate the images' essence and context with impressive precision. This paper delves into the details of this dataset's creation, methodology, and evaluation, underscoring its potential impact on the image captioning domain.

\subsection{Dataset}

The creation of a robust and diverse dataset is crucial for the success of any machine learning model, especially in the domain of image captioning. This section describes the process followed to produce the curated image captioning dataset, which subsequently played a cardinal role in enhancing the performance of our image captioning engine.
 
The initial step in our dataset creation involved the selection of images. From COCO's vast repository, a total of 10,000 images were randomly selected. This random selection ensured a broad representation,
        capturing the diversity and richness inherent in the COCO dataset while being compatible with the fine-tuning method and model used.

\subsection{Model-based Image Processing}

The dataset creation process involved a systematic approach in which, each image, sourced from the selected collection, was subjected to analysis using a series of advanced computer vision models. These models, each designed with specific functionalities, provided a range of interpretations and annotations for the images, ensuring a comprehensive understanding of the visual content.

The models employed in this analytical process are as follows:
\begin{itemize}
    \item \textbf{DETR} \citep{detr}: State-of-the-art object detection model based on the Transformer architecture. When processing an image, DETR identifies objects within the visual frame and provides annotations. These annotations include categorical labels that describe the objects and spatial coordinates that indicate their positions within the image. Named as \textit{OBJECTS MODEL 1} in the prompt.
    \item \textbf{Recognize Anything} \citep{ram}: This image tagging model has the capability to identify and tag a wide range of common categories within an image. The tags generated by this model serve as descriptors, offering a summarized representation of the image's content. Named as \textit{IMAGE TAGS} in the prompt.
    \item \textbf{OwLViT} \citep{owlvit}: OwLViT operates as an open-vocabulary object detection model. It takes a unique approach by using the outputs from the "Recognize Anything" model as its classes. For each tag produced by "Recognize Anything", OwLViT detects its presence and provides both its location and class name. This method complements the information provided by DETR, adding depth to the scene's understanding. Named as \textit{OBJECTS MODEL 2} in the prompt.
    \item \textbf{Tag2Text} \citep{tag2text}: Building upon the tags from "Recognize Anything", Tag2Text crafts captions that describe the image. This model's output provides a narrative that goes beyond simple object identification, offering a more holistic view of the image's content.  Named as \textit{SHORT CAPTION (Caption A)} in the prompt.
    \item \textbf{BLIP-2} \citep{blip2}: BLIP-2 is an image captioning model that, despite its reduced number of trainable parameters compared to some other models, has shown proficiency in its task. It provides detailed captions that describe the visual content of images. Named as \textit{SHORT CAPTION (Caption B)} in the prompt.
\end{itemize}

\begin{figure}[ht]
    \centering
    \includegraphics[width=1\linewidth]{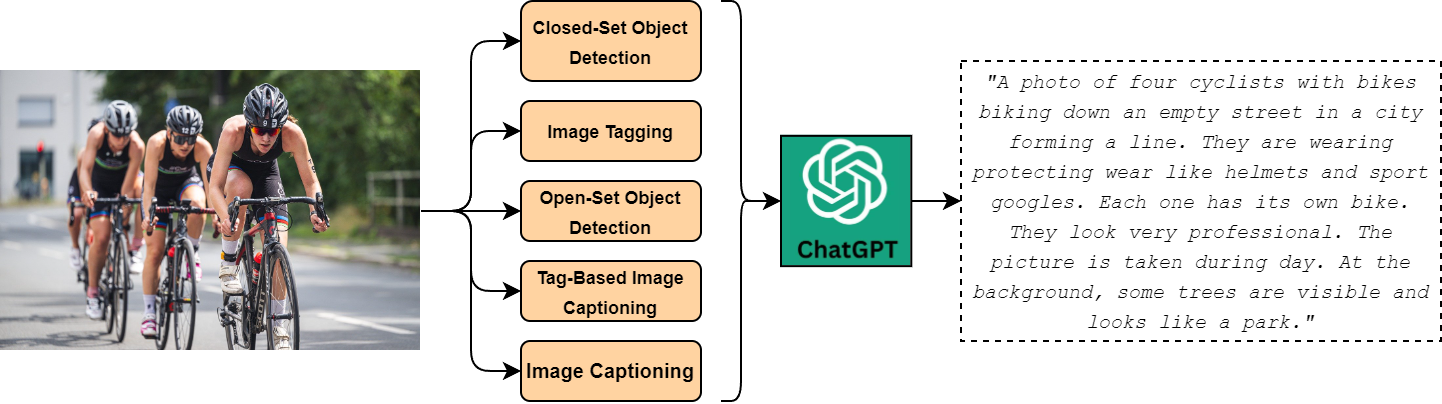}
    \caption{Each image goes through a variety of Computer Vision models that aid ChatGPT in the creation of very rich captions}
    \label{fig:image-pipeline}
\end{figure}

To consolidate the diverse outputs from these models, each was transformed into a text string. This string captures the essence of the information derived from each model's output. It is worth noting that, in this framework, ChatGPT interacts exclusively with these text strings, as it does not possess the ability to process images directly, thus becoming a kind of "imageless" image captioning. This process is described in Figure \ref{fig:image-pipeline}. The final prompt, which integrates the data from all the models and is then presented to ChatGPT, is presented in Figure \ref{fig:chatgpt-prompt}.

\begin{figure}[ht]
\noindent
\begin{minipage}{0.48\textwidth}
\begin{lstlisting}[basicstyle=\scriptsize\ttfamily]
# INPUTS
## IMAGE TAGS
'back', 'blanket', 'cat', 'control', 'couch', 'couple', 'curl', 'pad', 
'lay', 'pillow', 'pink', 'relax', 'remote', 'sleep', 'stretch', 'tabby'
## OBJECTS MODEL 1
cat at [344.06, 24.85, 640.34, 373.74]
remote at [328.13, 75.93, 372.81, 187.66]
remote at [39.34, 70.13, 175.56, 118.78]
cat at [15.36, 51.75, 316.89, 471.16]
couch at [-0.19, 0.71, 639.73, 474.17]
##OBJECTS MODEL 2
couch at [0.51, 1.19, 637.86, 479.38]
remote at [39.66, 72.18, 175.3, 117.2]
remote at [334.52, 76.81, 369.98, 182.76]
cat at [335.98, 14.93, 637.93, 373.28]
cat at [8.61, 54.8, 315.07, 479.78]
## SHORT CAPTION
Caption A: two cats laying on a couch with remote controls on the back
Caption B: a couple of cats on a couch with a remote
\end{lstlisting}
\end{minipage}
\hfill
\begin{minipage}{0.48\textwidth}
    \centering
    \includegraphics[width=0.8\linewidth]{example_3.jpg}
\end{minipage}
\caption{Joined prompt for ChatGPT from all the models mentioned. Each model corresponds to its section name output converted to plain text. The image on the right is the one used in the example. For reference, the output of ChatGPT for this example is: \textit{Two adorable cats are peacefully lounging on a comfortable couch, one on the left and the other on the right. They appear to be curled up and enjoying a restful moment. On the couch's backrest, you can see two remote controls neatly placed. The cats have distinctive tabby fur patterns, and their soft pink blankets are spread out beneath them. It seems like they are in complete control of their relaxation, creating a serene and cozy atmosphere.}}
\label{fig:chatgpt-prompt}
\end{figure}

\subsection{Fine-tuning}

The process of fine-tuning the BLIP-2 model with the novel dataset was a crucial step in achieving the results presented in this paper. This section delves into the specifics of how the model was adapted to this dataset, leveraging the strengths of both the BLIP-2 architecture and the LoRa method.

The dataset underwent preprocessing consistent with the methods employed in the original BLIP-2 paper \citep{blip2}. This ensured that the data was in an optimal format and structure, ready to be fed to the BLIP-2 model for fine-tuning.

For the fine-tuning process, the specific configuration of BLIP-2 used was BLIP-2 2.7B in 16-bit half-precision sourced from HuggingFace \citep{hfmodel}. This configuration was chosen for its balance between computational efficiency and performance capabilities.

The model was trained over a span of 5 epochs. This duration was chosen to prevent potential overfitting to the dataset. Given hardware constraints, a modest batch size of 3 was employed during training. The AdamW optimizer \citep{adamw} was utilized with a learning rate set to 5e-4. This low learning rate was specifically chosen to prevent catastrophic forgetting, ensuring that the model retained its pre-trained knowledge while adapting to the new dataset.

The LoRa method was seamlessly integrated into the fine-tuning process using the \texttt{peft} \citep{peft} library of adapters. The model, being sourced from HuggingFace, was loaded using the \texttt{transformers} \citep{transformerslib} library, making the integration straightforward. Specific hyperparameters associated with the LoRa method were adjusted to best suit our dataset:

\begin{itemize}
    \item \textbf{Rank} of the update matrices (\textit{r}): 16
    \item \textbf{Scaling} factor (\textit{lora\_alpha}): 16
    \item \textbf{Dropout} (\textit{lora\_dropout}): 0.1
    \item \textbf{Bias} training (\textit{bias}): "all"
    \item LoRa application \textbf{targets}: queries and keys (\textit{target\_modules})
\end{itemize}

The only modification made to the BLIP-2 model configuration was extending the \textit{max\_length} parameter in the tokenizer to 256. This adjustment was necessary to accommodate the richer text descriptions generated from the dataset.

\section{Evaluation}

For the evaluation of the proposed model, a new set of images that were not part of the training dataset was selected. These images were randomly sampled from the COCO dataset, ensuring that they were distinct from the 10,000 training images. Captions for these evaluation images were generated using the fine-tuned BLIP-2 model, with default parameters set at a \textit{temperature} of 1, \textit{top\_k} of 50, and \textit{top\_p} of 1.

To conduct a thorough evaluation of caption generation capabilities, the analysis was started by selecting a random set of 50 images from the test segment of the COCO dataset. Captions were then generated for these images using the PixLore and BLIP-2 models locally to ensure a consistent evaluation environment. For the GPT-4 and Google Bard models, their respective web applications were used, submitting each image with the standardized prompt: "Describe this image in a single paragraph." This prompt was crafted to be clear and unambiguous, minimizing potential bias from complex prompt-engineering, and specified the response to be in a single paragraph to align with our evaluation scope while avoiding the 256-token response limit, a constraint we considered sufficient for generating concise, yet informative captions.

The evaluation process was conducted with the help of 50 human evaluators, primarily from the general public, with a significant portion being Data Scientist fellows. More than 100 images were evaluated, ensuring a balance between layman perspectives and expert insights. 

\begin{figure}[ht]
  \centering
  \begin{minipage}{0.49\textwidth}
    \includegraphics[width=\linewidth]{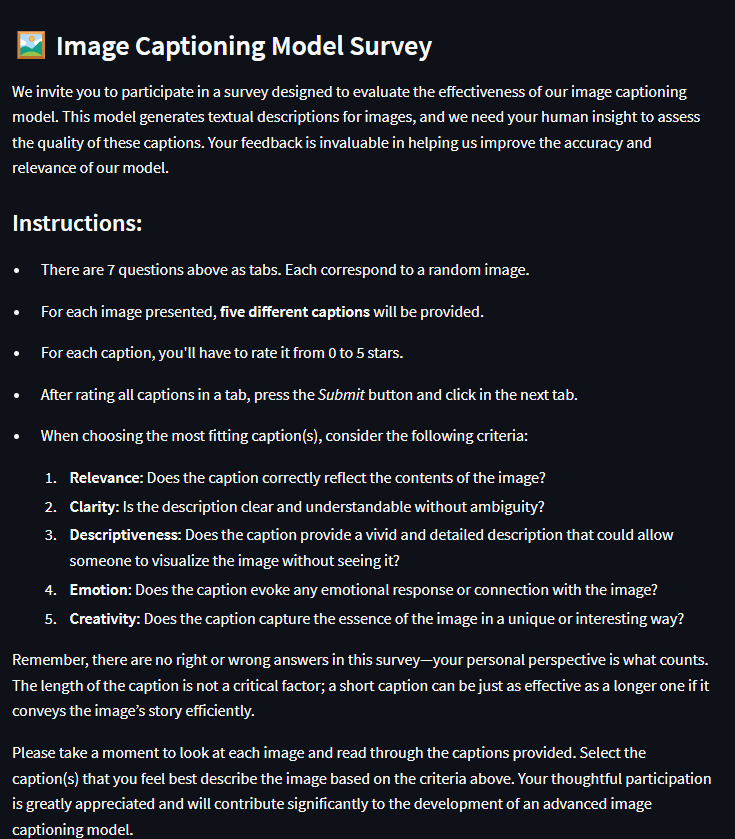}
  \end{minipage}
  \hfill
  \begin{minipage}{0.49\textwidth}
    \includegraphics[width=\linewidth]{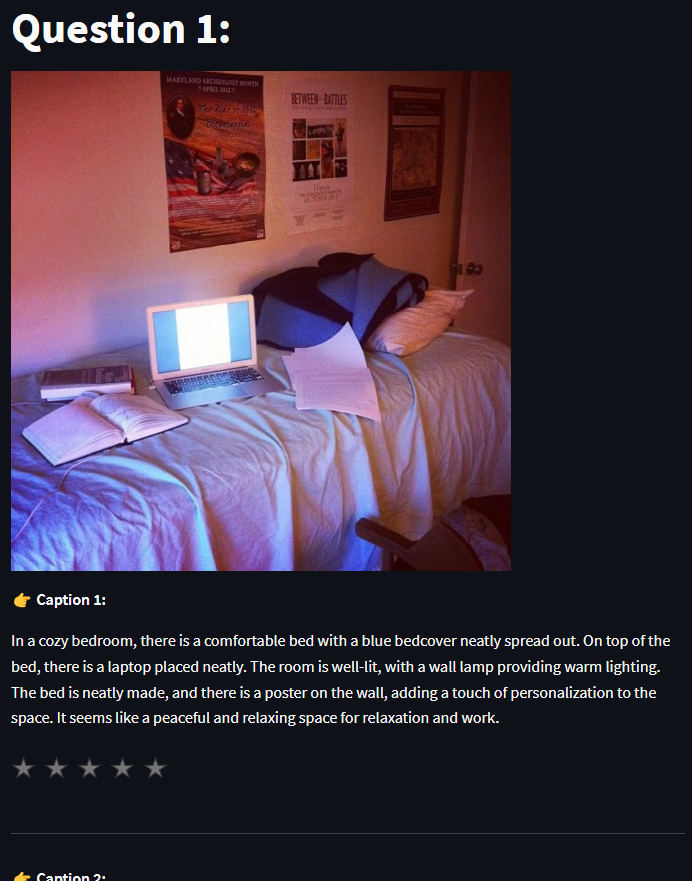}
  \end{minipage}
  \caption{Sample of the survey header, with the instructions given to the evaluators; and a sample of a question, with the five options of the models. The image is a random image of the test set of the COCO dataset and the captions are ordered randomly.}
  \label{fig:survey}
\end{figure}

A user-friendly survey was created using Streamlit for human evaluation. Evaluators were given specific guidelines to follow during the evaluation process. Specifically, they were shown three randomly selected images from the set, along with captions generated by Google Bard, GPT-4, BLIP-2 and PixLore, presented in random order. Participants rated each caption on a scale from 0 (lowest) to 5 (highest), allowing for independent evaluation of each caption's effectiveness based on criteria such as relevance, clarity, descriptiveness, emotion and creativity. Practical guidelines were provided to assist participants in their assessments, as depicted in Figure \ref{fig:survey}.

\section{Results}

The effectiveness of the generated captions was quantitatively analyzed by determining the percentage preference for captions from the proposed model, PixLore, over those from the other models. 

\begin{figure}
    \centering
    \includegraphics[width=0.8\linewidth]{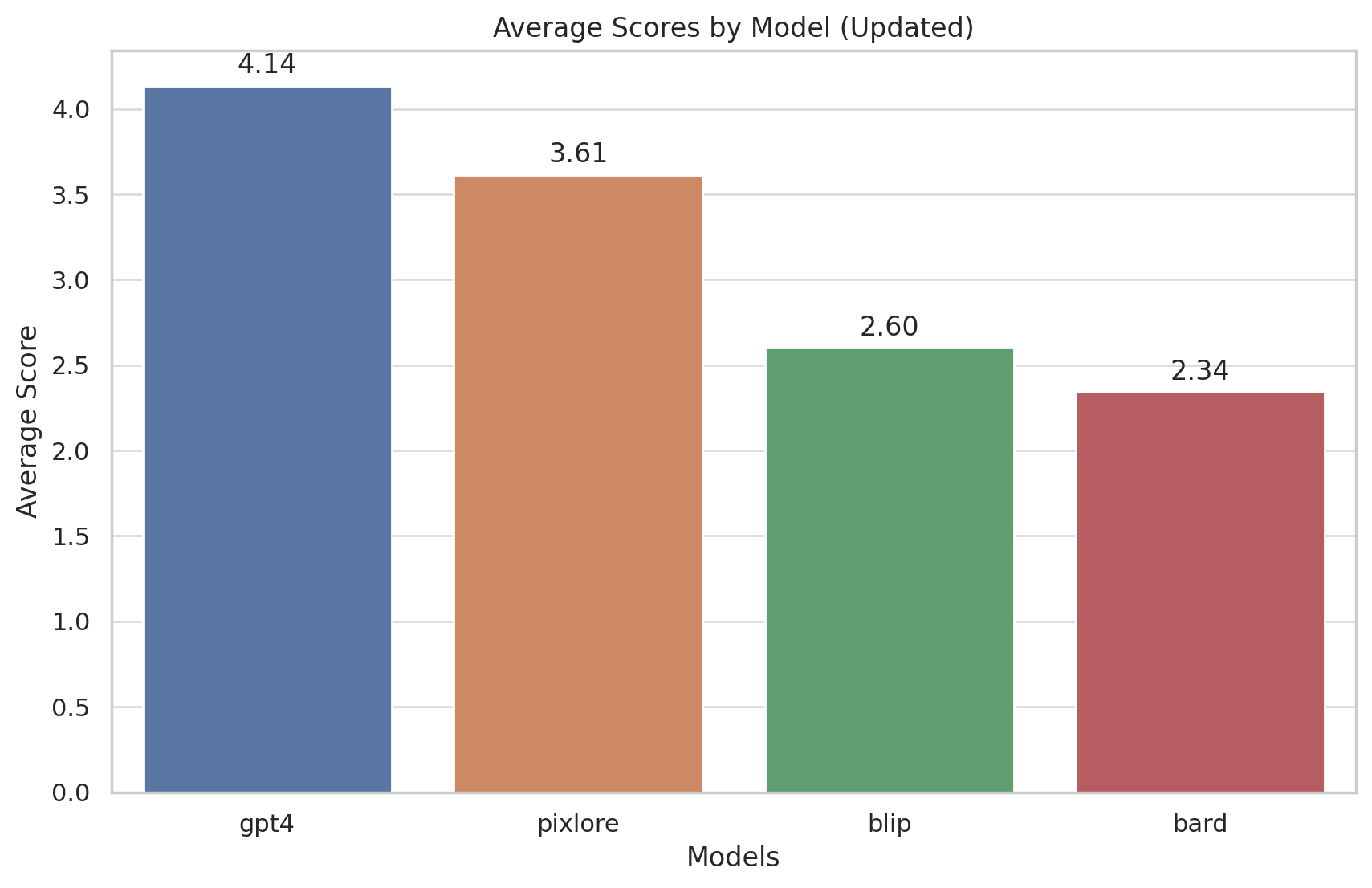}
    \caption{Comparison of average human evaluators' scores (0 to 5).}
    \label{fig:human-eval}
\end{figure}

Our analysis demonstrated a marked preference for PixLore's captions when compared to other models. Specifically, users preferred PixLore's captions over those from Bard and BLIP-2 in 69.32\% and 61.36\% of cases, respectively. Additionally, PixLore was preferred to GPT-4 in 26.14\% of the cases. This preference is visually represented in Figure \ref{fig:human-eval}, illustrating the human evaluation scores on a scale of 0 to 5.

Despite its smaller size, PixLore's performance was only marginally lower than GPT-4's by 12.80\%, yet it significantly outperformed Bard and BLIP-2, which were less preferred by approximately 35.18\% and 27.98\%, respectively. This was notable considering Bard's parameter count is roughly 50 times larger than PixLore's and BLIP-2's role as the baseline for our model, as observed in Table \ref{tab:model_parameters}.

\begin{table}[ht]
\centering
\caption{Average scores and parameter counts of each model. Parameters marked with an asterisk (*) are estimated.}
\begin{tabular}{lrr}
\toprule
\textbf{Model} & \textbf{Average Score} & \textbf{Parameters} \\
\midrule
\textbf{GPT-4} & 4.14 & 1.7T* \\
\textbf{PixLore} & 3.61 & 2.7B \\
\textbf{BLIP-2} & 2.60 & 2.7B \\
\textbf{Bard (Palm-2)} & 2.34 & 137B* \\
\bottomrule
\end{tabular}
\label{tab:model_parameters}
\end{table}

\begin{figure}[ht]
    \centering
    \includegraphics[width=0.8\linewidth]{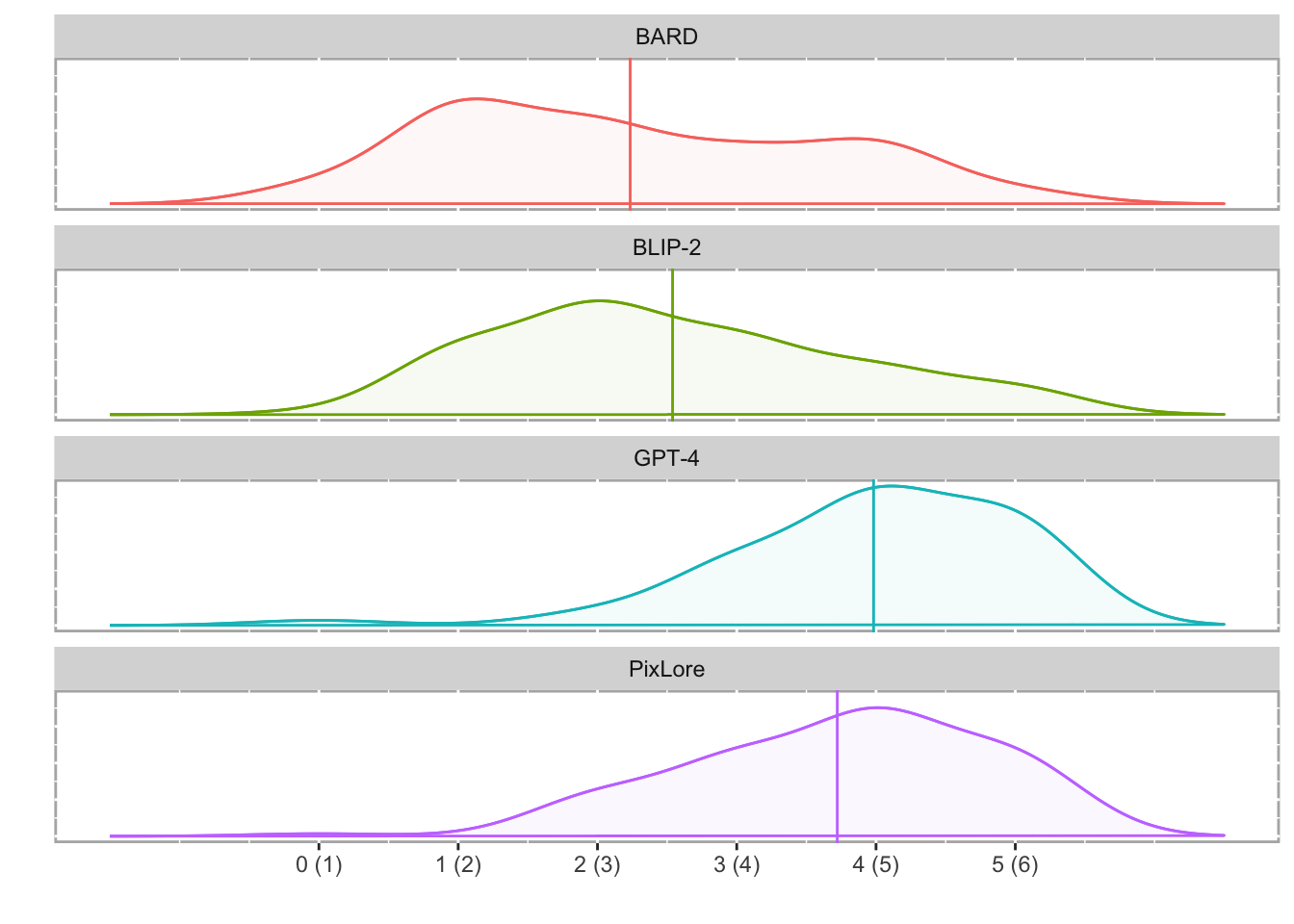}
    \caption{Density plots for each model's reponses.}
    \label{fig:density}
\end{figure}

\begin{figure}[ht]
    \centering
    \includegraphics[width=0.8\linewidth]{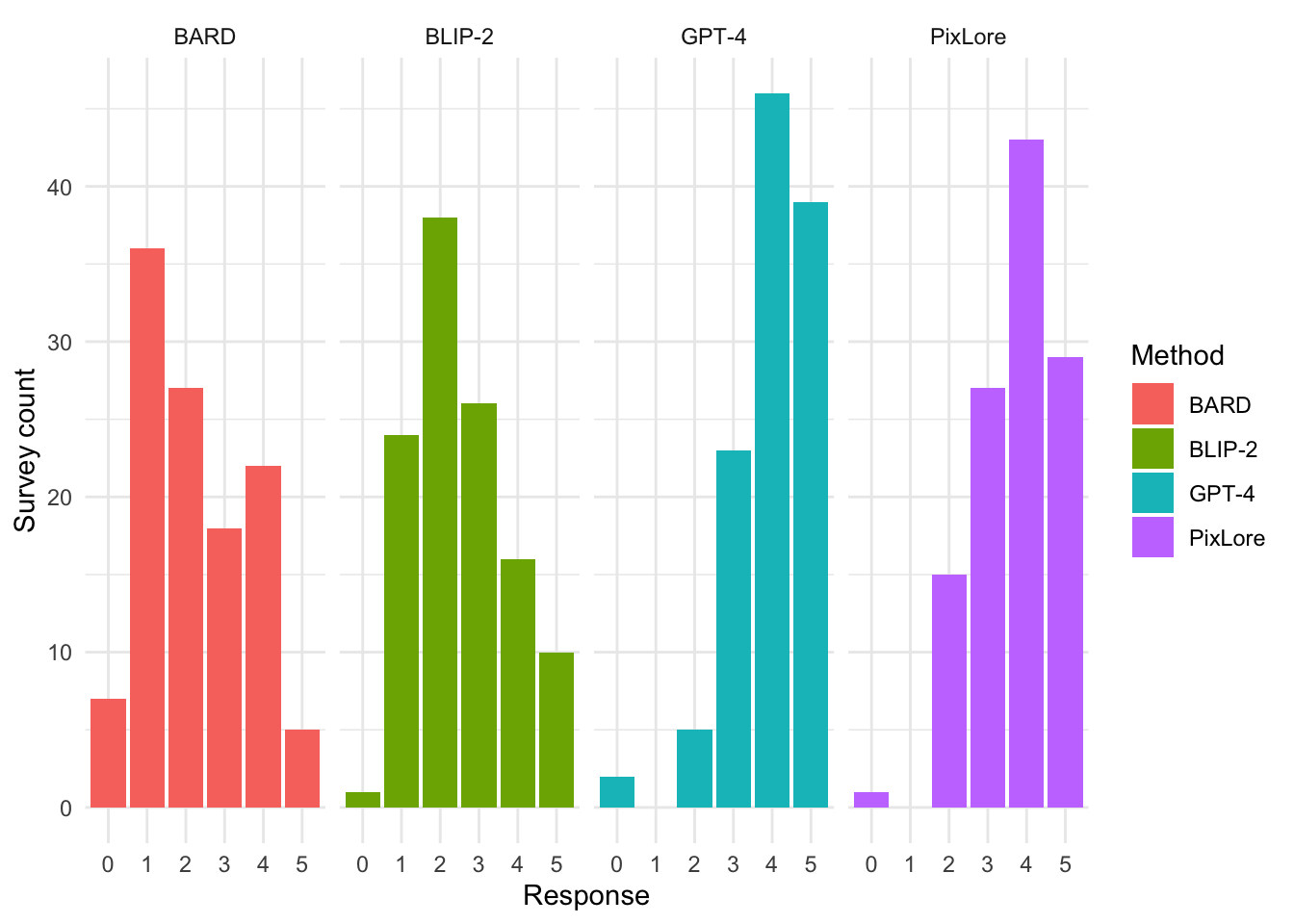}
    \caption{Histograms of the survey results per model.}
    \label{fig:ggplot_i}
\end{figure}

\begin{figure}[ht]
    \centering
    \includegraphics[width=0.8\linewidth]{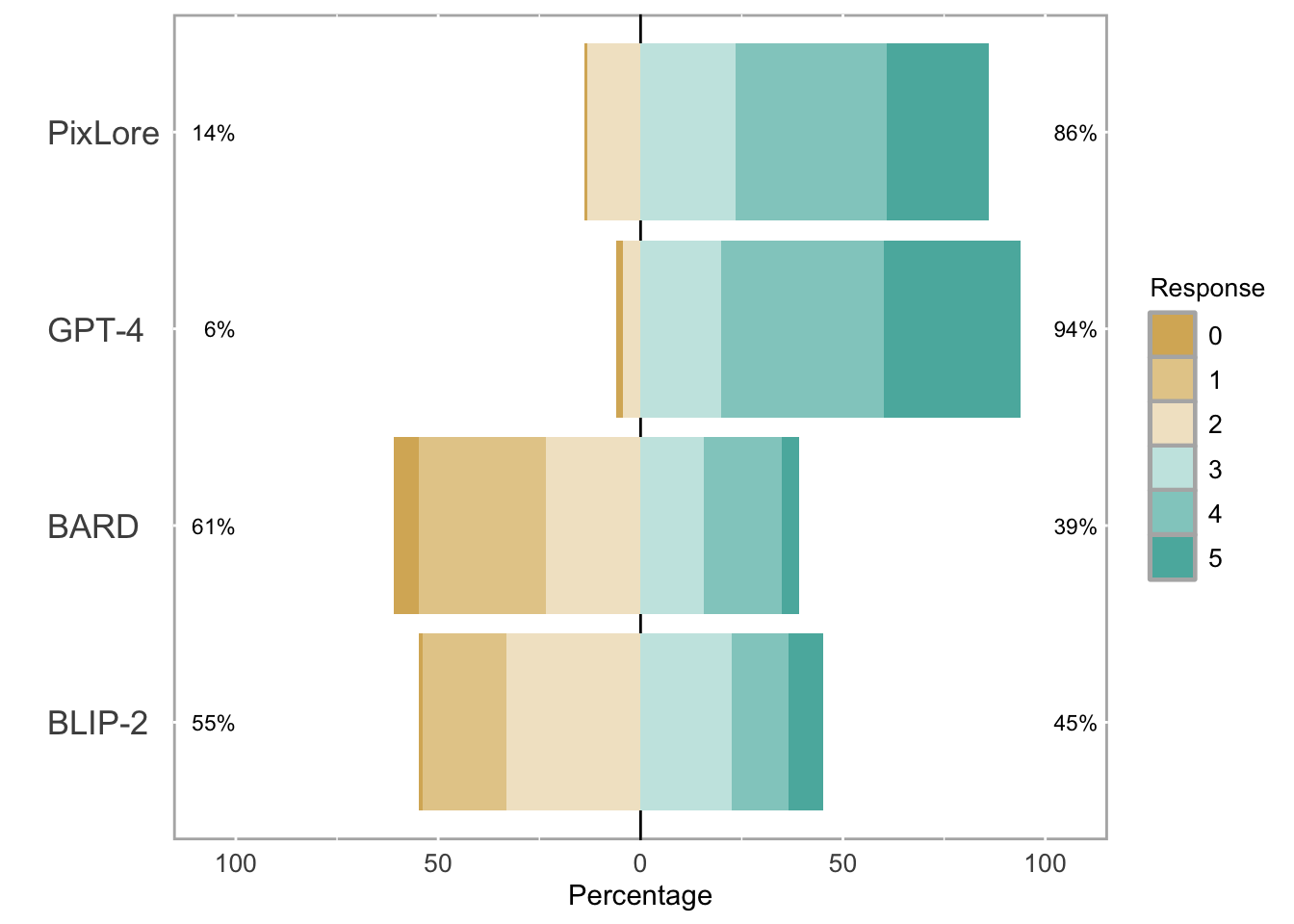}
    \caption{Diverging stacked bar chart with the proportion of evaluators who selected each response option.}
    \label{fig:likert}
\end{figure}

In a nuanced examination of survey results, PixLore emerges with distinct characteristics in user responses. The density plot in Figure \ref{fig:density} reveals a notable trend in the feedback, indicating that specific aspects of PixLore and GPT-4 resonate strongly with users. The bar plot in Figure \ref{fig:ggplot_i} further refines this perspective, indicating a pattern in the survey responses that sets these models apart from its counterparts. Complementing these findings, the diverging stacked bar chart in Figure \ref{fig:likert} illustrates the breadth of user reactions, offering a spectrum of insights into its reception. The combined graphical analyses provide a detailed view of user engagement with PixLore, revealing a complex but informative user experience.

For a more detailed understanding, some examples of the outputs generated by the different models are provided in Table \ref{table:examples}. This table illustrates the diversity in caption quality and style among the models, offering concrete examples to complement the quantitative results discussed earlier.

\begin{table}[ht]
\centering
\caption{Comparison of various state-of-the-art image captioning models versus PixLore.}
\vspace{5pt}
\resizebox{\textwidth}{!}{
\begin{tabular}{|p{3cm}|p{4cm}|p{4cm}|p{4cm}|}
    & \parbox[c]{1.\linewidth}{\centering\includegraphics[height=0.7\linewidth]{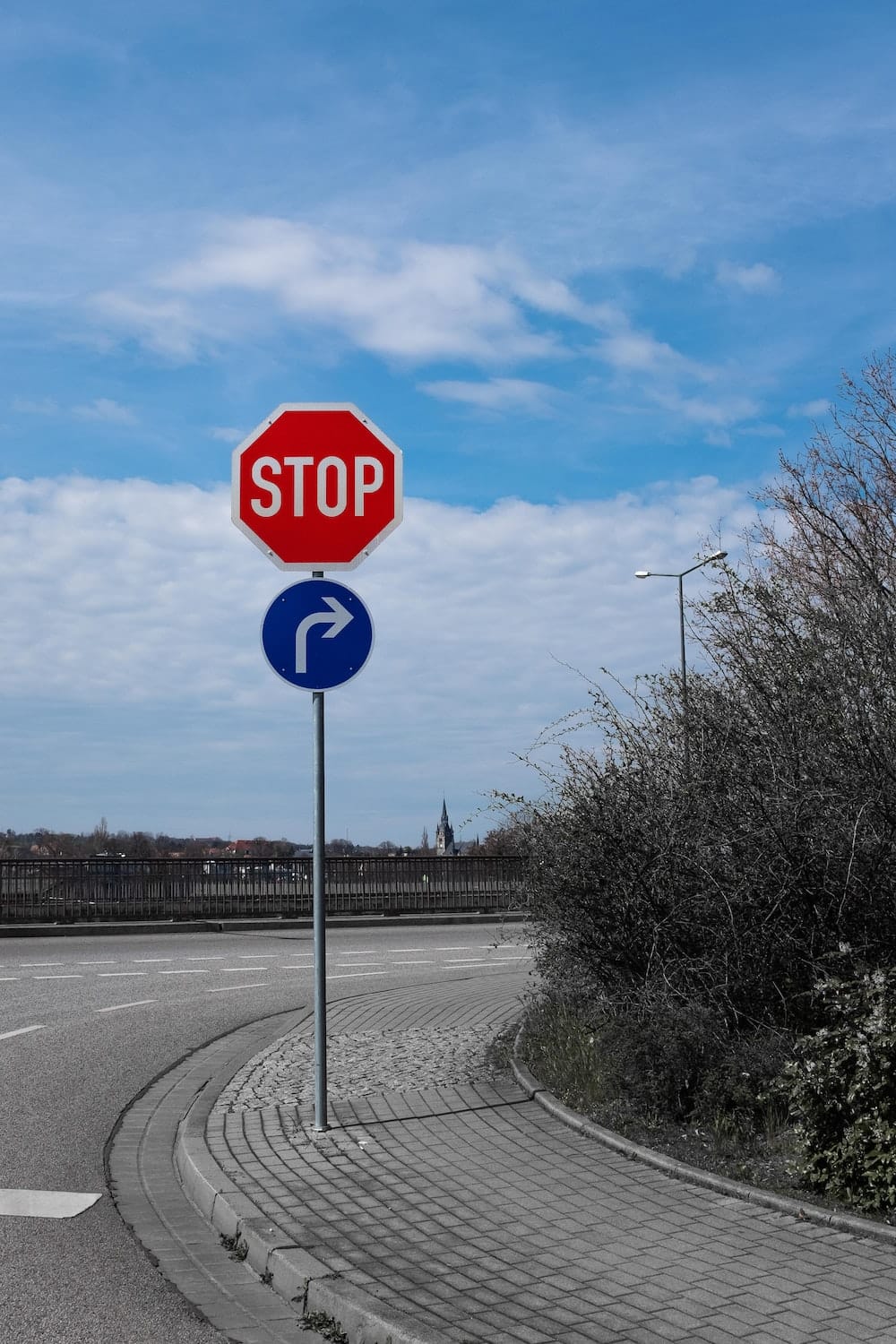}}
    & \parbox[c]{1.\linewidth}{\centering\includegraphics[height=0.7\linewidth]{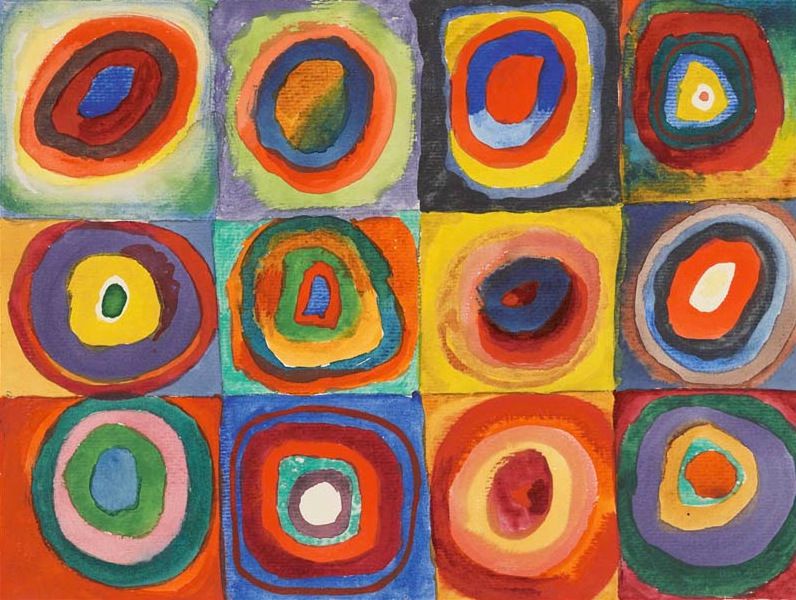}}
    & \parbox[c]{1.\linewidth}{\centering\includegraphics[height=0.7\linewidth]{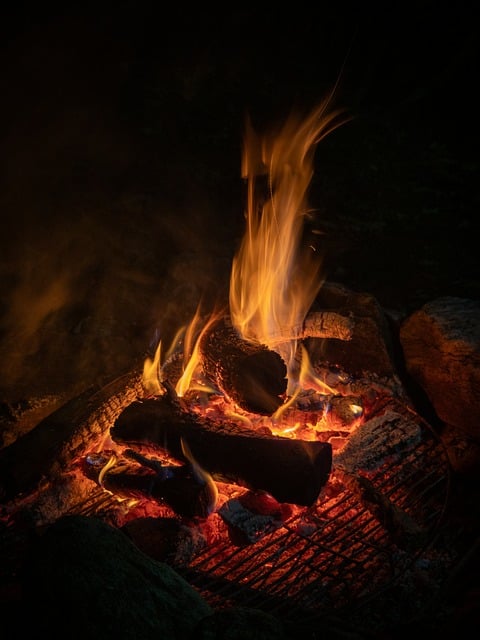}} \\\\
    
    \textbf{BLIP-2 Large \citep{blip2hf}}
    & \tiny there is a stop sign on the corner of a road.
    & \tiny a painting of a square with many different colors and shapes.
    & \tiny flames are lit in the dark near a fire pit. \\
    
    \textbf{Google Bard \citep{palm2})}
    & \tiny The image shows a red stop sign and a blue U-turn sign on a pole at a rural intersection. The stop sign is in the foreground and the U-turn sign is behind it. The signs are clearly visible and easy to identify. The image conveys the message that drivers must stop at the intersection and that U-turns are allowed.
    & \tiny This abstract painting features colorful circles arranged in squares. The bright, saturated colors and dynamic composition create a sense of joy and optimism. The painting is reminiscent of the work of Wassily Kandinsky, a pioneer of abstract art.
    & \tiny This close-up of a campfire at night captures the warmth and beauty of a gathering place. The flames reach up to the sky, casting a warm glow on the surrounding logs and rocks. The embers glow brightly, and the smoke rises in a thin wisp. The background is dark, with only a few stars visible. The image evokes the feeling of sitting around a campfire at night, surrounded by friends and family. It is a warm and inviting place to share stories and enjoy each other's company. \\
    
    \textbf{GPT-4 \citep{gpt4}}
    & \tiny This image shows a traffic sign that indicates a stop and a roundabout ahead. The sign is located on the side of a road with a pavement and some bushes. The sky is blue with some clouds and there are some buildings in the background. The image is realistic and has a lot of details.
    & \tiny This is an image of a colorful abstract painting that shows the artist's creativity and expression. The painting consists of 16 squares, each with a different color and pattern. The squares are arranged in a 4x4 grid, creating a contrast and harmony among them. The colors used are orange, red, blue, green, purple, and yellow, which are bright and vivid. The patterns are concentric circles, ovals, and rectangles, which are simple and geometric. The background is a light blue color, which creates a sense of space and calmness. The painting is done in a loose, brushy style, which adds texture and movement to the image.
    & \tiny The image shows a cozy campfire in the middle of a dark night. The fire is crackling and glowing with warm colors, creating a contrast with the blackness of the surroundings. The fire is built on a pile of rocks and logs, providing fuel and stability. The image also has a blurred square in the bottom right corner, which might be hiding some people who are enjoying the fire. The image captures the atmosphere of a relaxing and adventurous night outdoors. \\
    
    \textbf{PixLore}
    & \tiny A photo of a road with a stop sign on the side. The stop sign is positioned on the side of the road, indicating that it is a designated intersection. The sky above is cloudy, creating a gloomy atmosphere. The road is surrounded by a fence, and there is a curb separating the road from the adjacent area.
    & \tiny A colorful painting depicting a variety of circles and squares is displayed on a wall. The painting is vibrant and eye-catching, with a variety of colors and patterns. The circles and squares create a geometric pattern, adding a touch of creativity to the scene. The painting is a work of art, showcasing the artist's skill and creativity. The wall is covered in a soft, smooth surface, creating a comfortable and inviting atmosphere. The painting is a perfect addition to any room, adding a pop of color and creativity to the space.
    & \tiny A cozy campfire is lit up in the night, casting a warm glow on the surrounding area. The fire pit is filled with logs, creating a cozy and inviting atmosphere. The flames are burning brightly, adding to the ambiance of the scene. It seems like a perfect night for a campfire, with the smell of wood and the crackling of the logs filling the air. \\\\
\end{tabular}
}
\label{table:examples}
\end{table}

\section{Discussion and Future Work}

Image captioning models have recently seen a trend towards increasing model sizes, with parameter counts reaching into the billions and even trillions. For the pattern recognition community, this paper presents an alternative approach, showcasing a model that, despite having a significantly smaller parameter count, achieves competitive performance. This underscores the potential of targeted fine-tuning and the importance of a well-curated dataset. The model’s performance, when compared to models like Bard with 137 billion parameters and the speculated 1.7 trillion parameters of GPT-4, highlights the efficiency of the approach taken in this work. The reduced parameter count not only ensures computational efficiency but also demonstrates that state-of-the-art results can be achieved without the need for extremely large models.

The integration of outputs from various computer vision models using ChatGPT offers a novel approach to image captioning. This work leverages ChatGPT’s capability to understand tasks and the inputs provided, effectively acting as an indirect form of knowledge distillation. Instead of directly extracting knowledge from teacher models, the model assimilates information from various state-of-the-art computer vision models, transforming its capabilities in image captioning. The detailed descriptions generated by the model hint at the latent potential of its image encoder. While the specific applications for scene understanding have not been delineated in this work, the encoder’s ability to generate detailed and nuanced descriptions suggests its potential utility in applications requiring detailed scene interpretations, such as real-time scene analysis or augmented reality systems.

The "no free lunch theorem" in machine learning posits that no single model can be universally optimal across all tasks. This work provides empirical evidence supporting this theorem. A model fine-tuned for a specific task, especially when trained on a rich curated dataset, can potentially outperform more general models. The dataset-driven approach adopted in this work not only achieves commendable performance but does so with a significantly reduced parameter count, leading to benefits like reduced computational costs and faster inference times.

\section*{Conclusion}

The encouraging outcomes illustrated by our image captioning model present compelling opportunities for future investigation, particularly within the domain of scene comprehension tasks in the broader pattern recognition domain. The sophisticated capabilities of our vision encoder in fine-grained object recognition and comprehensive scene understanding suggest its potential as a robust feature extractor for complex scene analysis applications. Future work could explore the integration of our model within systems that require a deeper semantic understanding of visual inputs, such as autonomous navigation systems where accurate environmental interpretation is crucial, or in intelligent surveillance systems where distinguishing subtle nuances could be paramount for safety and security.

Additionally, the use of our richly annotated caption dataset offers a fertile ground for the pre-training of image-text models like CLIP \citep{clip}. By leveraging our dataset, which has been proven to enhance captioning quality, we could augment the richness of the features that such models learn. This could, in turn, enhance their performance across a wide range of vision-and-language tasks. Furthermore, the extension of the token context window from the standard 77 tokens to 256 tokens, as used in our work, suggests an interesting research direction. This increased context window could allow for the capturing of more complex relationships and subtleties within scenes, potentially leading to a more nuanced and comprehensive understanding of visual content.

Moreover, the fine-tuning of text-to-image models using our approach presents a tantalizing opportunity to improve the generation of detailed images from longer, more complex descriptions. By feeding these models with fine-grained, descriptive captions, they could learn to render images that better reflect the intricacies of the input text, thus enhancing the synthesis of rich visual scenes. This could significantly benefit domains such as digital art and design, where the ability to generate detailed visual content from descriptive language can streamline creative workflows.

In conclusion, this paper emphasizes the significance of fine-tuning, the innovative approach of using ChatGPT for indirect knowledge distillation, and the paramount importance of a rich curated dataset in the domain of image captioning and pattern recognition. As research in this field progresses, this work provides a foundation, emphasizing the balance between model efficiency, innovation, and data quality.


\bibliography{references}

\end{document}